\documentclass{article}

\PassOptionsToPackage{numbers, compress}{natbib}

    \usepackage[preprint]{neurips_2024}

\usepackage[utf8]{inputenc} 
\usepackage[T1]{fontenc}    
\usepackage{hyperref}       
\usepackage{url}            
\usepackage{booktabs}       
\usepackage{amsfonts}       
\usepackage{nicefrac}       
\usepackage{microtype}      
\usepackage{xcolor}         
\usepackage{amsmath} 
\usepackage{algorithm}
\usepackage{multirow}
\usepackage{algpseudocode}
\usepackage{graphicx}
\newtheorem{theorem}{Theorem}
\usepackage{wrapfig}

\title{Certifying Adapters: Enabling and Enhancing the Certification of Classifier Adversarial Robustness}
\author{%
  Jieren Deng \\
  University of Connecticut \\
  \texttt{jieren.deng@uconn.edu} \\
  \And
  Hanbin Hong \\
  University of Connecticut \\
  \texttt{hanbin.hong@uconn.edu} \\
  \And
  Aaron Palmer \\
  University of Connecticut \\
  \texttt{aaron.palmer@uconn.edu} \\
  \And
  Xin Zhou \\
  Baidu USA \\
  \texttt{zhouxin16@baidu.com} \\
  \And
  Jinbo Bi \\
  University of Connecticut \\
  \texttt{jinbo.bi@uconn.edu} \\
  \And
  Kaleel Mahmood \\
  University of Connecticut \\
  \texttt{kaleel.mahmood@uconn.edu} \\
  \And
  Yuan Hong \\
  University of Connecticut \\
  \texttt{yuan.hong@uconn.edu} \\
  \And
  Derek Aguiar \\
  University of Connecticut \\
  \texttt{derek.aguiar@uconn.edu} \\
}

\begin{document}

\maketitle

\begin{abstract}
Randomized smoothing has become a leading method for achieving certified robustness in deep classifiers against $\ell_p$-norm adversarial perturbations. 
Current approaches for achieving certified robustness, like data augmentation with Gaussian noise and adversarial training, require expensive training procedures that tune large models for different Gaussian noise levels from scratch and thus cannot leverage high-performance pre-trained neural networks.  
In this work, we introduce a novel \textit{certifying adapters framework} (CAF) that enables and enhances the certification of classifier adversarial robustness. 
Our approach makes few assumptions about the underlying training algorithm or feature extractor, and is thus broadly applicable to different feature extractor architectures (e.g., convolutional neural networks or vision transformers) and smoothing algorithms. 
We show that CAF (a) enables certification in uncertified models pre-trained on clean datasets and (b) substantially improves the performance of certified classifiers via randomized smoothing and SmoothAdv at multiple radii in CIFAR-10 and ImageNet. 
We demonstrate that CAF achieves improved certified accuracies when compared to methods based on random or denoised smoothing, and that CAF is insensitive to certifying adapter hyperparameters.
Finally, we show that an ensemble of adapters enables a single pre-trained feature extractor to defend against a range of noise perturbation scales.

\end{abstract}

\section{Introduction}

Deep learning classifiers based on vision transformer (ViT)~\cite{dosovitskiy2020vit} and convolutional neural network (CNN)~\cite{He2015DeepRL} architectures have demonstrated high accuracy on in-distribution images, but are vulnerable to adversarial examples~\cite{szegedy2013intriguing, goodfellow2014explaining}. 
In the context of image classification, adversarial examples are inputs with imperceptible perturbations that are designed to increase the likelihood of misclassification~\cite{szegedy2013intriguing}. 
In response to these threats, many empirical defenses have been proposed that increase classifier robustness to adversarial examples~\cite{BaRT, TiT, wang2023better}, but do not provide theoretical guarantees and are often broken by specialized adaptive attacks~\cite{tramer2020adaptive, mahmood2021beware, sitawarin2022demystifying, rathbun2022game}. 
In contrast, \textit{certified defenses} guarantee robustness to adversarial examples within a specified class~\cite{Raghunathan2018CertifiedDA,wong2018provable}, though early certification methods had difficulties scaling to large networks and datasets~\cite{rs}.

The first certified classifier that extended to ImageNet was trained using \textit{randomized smoothing}~\cite{Lcuyer2018CertifiedRT}.
Randomized smoothing creates a smoothed classifier that is certifiably robust to adversarial perturbations under a specified $\ell_p$-norm (e.g., $\ell_1$~\cite{Lcuyer2018CertifiedRT}, $\ell_2$~\cite{rs}, or $\ell_\infty$~\cite{yang2020randomized}) by corrupting training examples with noise sampled from an assumed distribution (usually Gaussian)~\cite{10.5555/3524938.3525929}.
One limitation is that the radius with which smooth neural network classifiers are adversarially robust cannot be computed exactly~\cite{rs}.
However, recent work shows the radius can be approximated with arbitrarily high probability using Monte Carlo samples of perturbed examples followed by classification by majority vote~\cite{rs}. 

\begin{figure}
\centering
\includegraphics[width=1\textwidth]{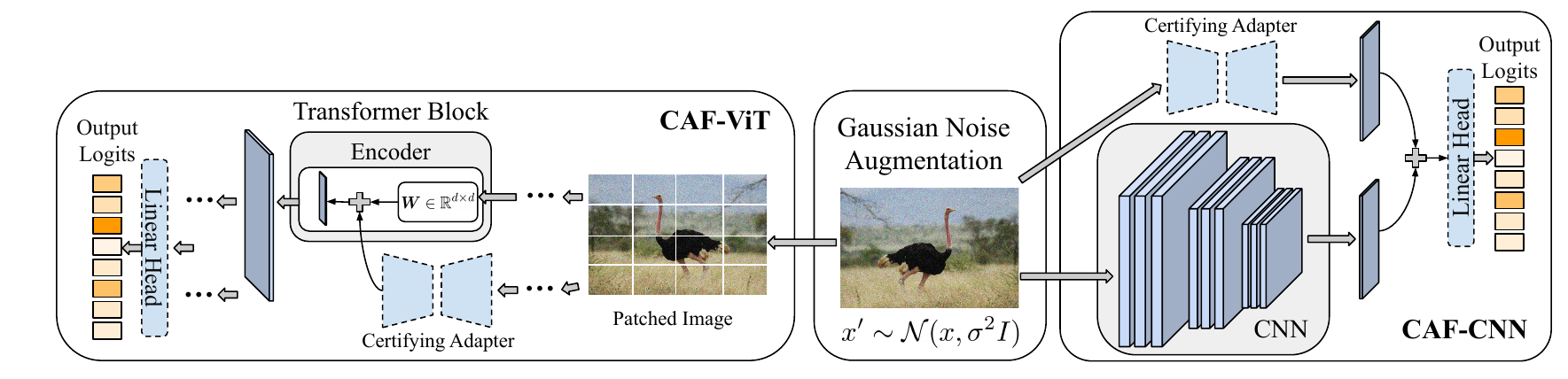}
\caption{\textbf{The certifying adapter framework for a single input example.}
Certifying adapters are defined for both ViT (CAF-ViT) and CNN (CAF-CNN) architectures. 
Trainable model components are denoted with a dashed outline. \( W \in \mathbb{R}^{d \times d} \) represents the weight matrix used in the encoder part of the Transformer Block.
}
\label{Fig:intro_fig}
\end{figure}

While randomized smoothing-based certification is the only verification approach to extend to ImageNet-scale data~\cite{sok_crdnn}, the smoothed classifier requires training for each noise-augmented data distribution.
Training with a smaller noise scale often leads to better certified accuracy but within a smaller robustness radius, whereas a larger noise scale usually achieves a larger radius. 
Additionally, training with noise-augmented data typically results in diminished performance on clean (unperturbed) data~\cite{zhang2019theoretically}. 
As the number of high-performance pre-trained models increases, generating certified smoothed classifiers for various machine learning tasks, and evaluating the trade-offs between clean accuracies, certified accuracies, noise scale, and robustness radius become prohibitively expensive. 

\begin{wrapfigure}{r}{0.35\textwidth}
\centering
\includegraphics[width=0.35\columnwidth]{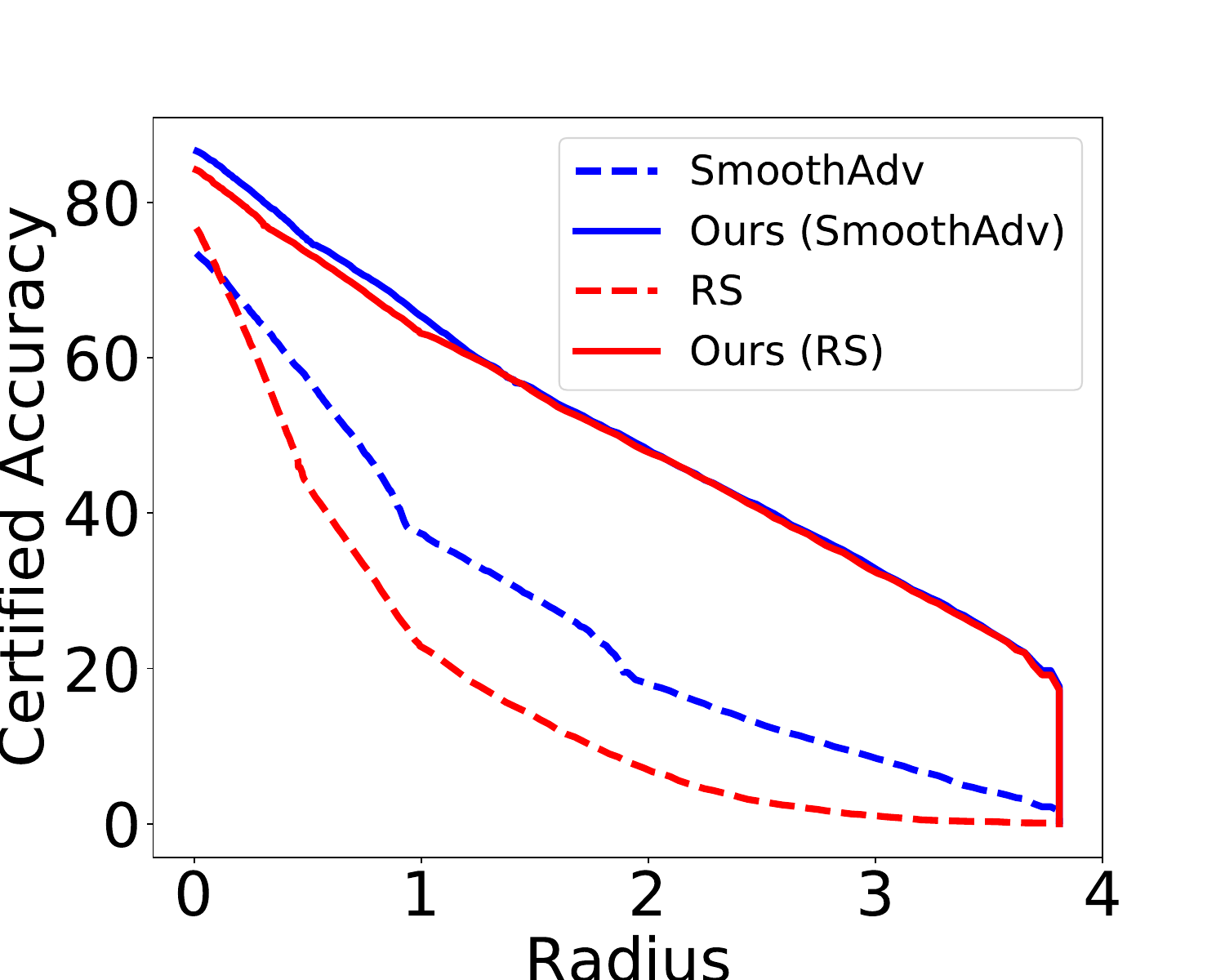}
\caption{\textbf{Upper envelope of certified accuracy.} 
CAF (Ours) is compared with RS~\cite{rs} and SmoothAdv~\cite{smoothadv} on CIFAR-10.
}
\label{Fig:upper_evelop}
\end{wrapfigure}


Motivated by the effectiveness of lightweight adapters in enhancing the performance of pre-trained models~\cite{hu2022lora}, this work seeks to answer two primary research questions: \textbf{RQ1}: can uncertified image classifiers be adapted to provide certified adversarial robustness?
\textbf{RQ2}: to what extent, if any at all, can adapters improve the certified robustness of certifiably robust image classifiers?
We answer both of these questions in the affirmative with a new framework composed of (1) a pre-trained feature extractor, (2) a  linear prediction head, and (3) a \textit{certifying adapter} (Fig.~\ref{Fig:intro_fig}).
In brief, the latent representations of an input image $x$ are generated using a pre-trained feature extractor and a trainable \textit{certifying adapter}; these representations are combined and classified using a linear prediction head. 
In our setup, the feature extractor is derived from a pre-trained CNN or ViT classification model, which can be certified or uncertified. 
Our certifying adapter framework (CAF) is generally applicable to a large variety of models and tasks and does not require retraining large models.
We show that CAF achieves state-of-the-art certified accuracies across all radii in CIFAR-10 and competitive performance in ImageNet compared to randomized smoothing-base classifiers and diffusion-based purification methods for an uncertified ViT (\textbf{RQ1}). 
With regards to \textbf{RQ2}, CAF shows a notable improvement in certified accuracy across radii in $[0,4]$ on the CIFAR-10 dataset when compared to the certified classification models of the randomized smoothing (RS) method in Cohen \textit{et. al.}~\cite{rs} and SmoothAdv~\cite{smoothadv}.
In summary, our contributions include:
\begin{itemize}
\item We propose the certifying adapter framework (CAF), which is capable of adapting the latent representation of both ViT- and CNN-based pre-trained feature extractors.
\item We show that CAF (a) enables the robustness certification of uncertified models trained on benchmark (clean) datasets, and (b) enhances the performance of state-of-the-art certified models at various radii on both CIFAR-10 and ImageNet. 
\item We demonstrate that an ensemble of certifying adapters trained at different noise scales enables a single pre-trained feature extractor to defend against noise perturbations at multiple scales.
\item Finally, ablation studies show that the certified accuracies achieved by CAF are insensitive to the certifying adapter size or the adapter rank.
\end{itemize}

\section{Preliminaries}
\textbf{Certified Robustness. } Given a radius $r \in \mathbb{R}^{+}$, certified robustness provides theoretical guarantees on the maximum perturbation of an input such that the model prediction remains the same~\cite{sok_crdnn}.
Formally, robustness certification for a classifier $f(\cdot)$ ensures that given a clean data example $x$ and an adversarial example $x'$, the condition $\arg\max_c f(x) =\arg\max_c f(x')$ holds when the distance between $x$ and $x'$ satisfies $d(x,x') < r$, where $d(\cdot)$ is some norm in the set $\{\ell_1, \ell_2, \dots, \ell_{\infty} \}$ and $c \in \mathcal{Y}$ with $\mathcal{Y}=\{1,\dots,C\}$ denoting the class labels.

\noindent \textbf{Randomized Smoothing.} 
Randomized smoothing (RS) is an algorithm for certifying robustness for a smoothed classifier $G_R$ based on an arbitrary classifier \( f \) by perturbing inputs with Gaussian noise~\cite{rs}: 
\begin{equation*}
G_{R}(x) = \arg\max_{c \in \mathcal{Y}} \mathbb{P}(f(x + \epsilon) = c), \quad \epsilon \sim \mathcal{N}(0, \sigma^2I)
\end{equation*}
where \(\mathcal{N}(0, \sigma^2I)\) represents i.i.d. Gaussian noise with standard deviation $\sigma$.
Given an input \(x\), the prediction of the smoothed classifier is the most likely class over the noisy input, $(x + \epsilon)$.
The noise level, denoted as $\sigma$, controls the trade-off between robustness against perturbations and clean accuracy. 
As $\sigma$ increases, the robustness of the smoothed classifier increases, but at the cost of a decrease in clean accuracy.
Let $p_{A} = \mathbb{P}(f(x+\epsilon) = c_{A})$ and $p_{B} = \max_{c_{B} \ne c_{A}}(\mathbb{P}(f(x+\epsilon) = c_{B})$ be the probability associated with the most probable and ``runner-up'' classes of the smoothed classifier $G_{R}(\cdot)$, respectively. 
Then, the smoothed classifier $G_{R}(\cdot)$ is provably robust to an adversarial example $x_{adv}$ around $x$ with a radius defined as $ ||\delta||_2 \equiv ||x_{adv} - x||_2 \leq r$, where $r = \frac{\sigma}{2} \left( \Phi^{-1}(p_A) - \Phi^{-1}(p_B) \right)$ and $\Phi^{-1}$ denotes the inverse of the standard Gaussian cumulative distribution function. 
To achieve certified robustness, RS~\cite{rs} trains a smoothed classifier using data augmented with Gaussian noise, whereas SmoothAdv~\cite{smoothadv} uses adversarial training techniques such as projected gradient descent~\cite{madry2018towards}.


\noindent \textbf{Denoised Smoothing.} 
Although RS is the only verification approach that can certify classifiers on ImageNet-scale data~\cite{sok_crdnn}, it requires training a smoothed classifier from scratch for each noise scale, which can be computationally prohibitive.
Denoised smoothing methods address this bottleneck by using a denoising model $\phi(\cdot)$ to remove image perturbations from an adversarial example $x_{adv}$ without training a classifier from scratch~\cite{nie2022diffusion, yoon2021adversarial, allen2022feature,xiao2022densepure,diffsmooth, creswell2018denoising,wu2023dmae,hwang2019puvae}.
Then, a smoothed denoising classifier, $G_{D}(\cdot)$, can be formulated as:
\begin{equation*}
G_{D}(x) = \arg\max_{c \in \mathcal{Y}} \mathbb{P}(f(\phi( x + \epsilon)) = c), \quad \epsilon \sim \mathcal{N}(0, \sigma^2I)
\end{equation*}
There are two primary methods of denoising: denoising autoencoders~\cite{creswell2018denoising,wu2023dmae,hwang2019puvae} and diffusion-based models~\cite{nie2022diffusion, yoon2021adversarial, allen2022feature,xiao2022densepure,diffsmooth}. 
DMAE~\cite{wu2023dmae} uses a masked autoencoder~\cite{9879206} that is pre-trained on a large dataset (e.g., ImageNet), followed by additional pre-training with noise-augmented data. 
Subsequently, a linear head is added to the encoder component for classifier fine-tuning.
In contrast, diffusion-based approaches typically use of off-the-shelf diffusion models for purification. 
DiffSmooth generates noisy input as in RS, but then purifies each image and adds a second round of noise to each purified image for certified classification~\cite{diffsmooth}. 
Consequently, DiffSmooth requires a multiplicative factor more noisy image generation than RS and is also sensitive to the time-step used for purification~\cite{carlini2023free}.

\section{The Certifying Adapter Framework}
In this section, we introduce the certifying adapter framework (CAF), which eliminates the need for training a smoothed classifier from scratch by adapting a pre-trained feature extractor (CNNs or ViTs) to enable robustness certification of uncertified models and improve the performance of certified models.
We additionally describe an ensemble of certifying adapters, which enables a single pre-trained feature extractor to achieve certified robustness against multi-scale noise perturbations using hierarchical adaptation.

\subsection{Pre-trained Feature Extractor and Linear Predictor}
We denote the base classifier as the parameterized model $f_{\theta
}(\cdot)$. 
The base classifier can be further described as having two main components: a feature extractor, represented as $f_{\theta_z}(\cdot)$, and a linear head predictor, denoted as $f_{\theta_y}(\cdot)$. 
The latent representation, $z$, of a given input $x$ is output by applying $f_{\theta_z}(\cdot)$ as in $f_{\theta_z}(x)$, where $f_{\theta_z}(\cdot)$ is constructed using CNNs or ViTs. 
The linear predictor $f_{\theta_y}(\cdot)$ consists of one or more fully connected layers, which generates a vector of logits and a class prediction with maximum probability $\hat{y}$.
\begin{align*}
    f_{\theta}(x) = f_{\theta_y}(f_{\theta_z}(x)), \quad f_{\theta_z}(x) = z , \quad \arg\max f_{\theta_y}(z) = \hat{y}
\end{align*}
\subsection{Certifying Adapter}
Assume a latent representation \( z_y \) that is located in a specified latent space \( \textbf{Z}_{Y} \) is deterministically classified as $y$ by the linear head predictor $f_{\theta_y}(\cdot)$, where $y$ is the class label for the corresponding $x$.
\begin{equation*}
\arg\max f_{\theta_y}(z_y) \equiv y, \quad z_y \in \textbf{Z}_{Y}
\end{equation*}
Given the deterministic pre-trained feature extractor $f_{\theta_z}(\cdot)$, the latent representation $z'$ derived from noisy input data $x'$ may not necessarily fall within this desired space, especially if the feature extractor has not been pre-trained with data augmented by noise.

In our framework, we specify a \textit{certifying adapter}, $f_{\theta_a}(\cdot)$, which receives noised example $x'$ as input and produce a latent representation denoted as $z_a$.
This representation $z_a$ is then added with the latent representation from the pre-trained feature extractor, $z'$, resulting in $z'+z_a$.
The goal is to optimize $z'+z_a$ to fit appropriately within the target space \( \textbf{Z}_{Y} \).
However, since \( \textbf{Z}_{Y} \) is an ideal yet undefined space, our strategy aims to approximate $z'+z_a$ to $z_y$ by optimizing $\arg\max f_{\theta_y}(z' + z_a) \rightarrow y$, where $f_{\theta_z}(x') = z'$, and $f_{\theta_a}(x') = z_a$.


\subsection{Smoothed Adaptive Classifier}
We integrate the pre-trained feature extractor $f_{\theta_z}(\cdot)$, linear predictor $f_{\theta_y}(\cdot)$, and the certifying adapter $f_{\theta_a}(\cdot)$ as the \textit{adaptive} classifier:
\begin{equation*}
F(x) = f_{\theta_y}(f_{\theta_z}(x) + f_{\theta_a}(x))
\end{equation*}
With this adaptive classifier, $F(\cdot)$, and a labeled data instance $(x, y)$, we define the smoothed adaptive classifier $G(\cdot)$ that returns the class $y$ of a set of noised images sampled from $\mathcal{N}(x, \sigma^2I)$ as: 
\begin{align*}
G(x) &= \arg\max_{y \in \mathcal{Y}} \mathbb{P}[F(x + \epsilon) = y]\\
&= \arg\max_{y \in \mathcal{Y}} \mathbb{P}[f_{\theta_y}[f_{\theta_z}(x + \epsilon) + f_{\theta_a}(x + \epsilon)] = y] \quad \text{with} \quad \epsilon \sim \mathcal{N}(0, \sigma^2I).
\end{align*}
We leverage RS to provide the following certified robustness guarantee for $G(\cdot)$~\cite{rs}.
\begin{theorem} (Adapted from Cohen \textit{et. al.}~\cite{rs}).
    Given the above defined $F(\cdot)$, $G(\cdot)$ and $(x, y)$, assume that $G(\cdot)$ correctly classifies $x$ as $y$. Then, there exists a radius $r$ such that for any $x'$ satisfying the condition $||x' - x||_p \leq r$, it holds that $G(x') = G(x)$. Furthermore, $r$ can be approximated as:
    \begin{equation*}
        r = \frac{\sigma}{2} [\Phi^{-1}(\mathbb{P}[f_{\theta_y}(f_{\theta_z}(x + \epsilon) + f_{\theta_a}(x + \epsilon)) = y]) - \Phi^{-1}(\max \mathbb{P}[f_{\theta_y}(f_{\theta_z}(x + \epsilon) + f_{\theta_a}(x + \epsilon)) \neq y])]
    \end{equation*}
    where \( \Phi^{-1} \) represents the inverse of the standard Gaussian cumulative distribution function. 
\end{theorem}
Since $r$ cannot be computed analytically, we use $f_{\theta_{y}} (\cdot)$ to calculate the two classes with the highest probability from a set of randomly noised images $\{x'\}$ and then use their empirical probabilities to approximate the certified radius. 
The noise scale $\sigma$ is a hyperparameter that controls the degree of perturbation applied to $x$.

We train $F$ by maximizing the number of noised inputs that are correctly classified~\cite{Lcuyer2018CertifiedRT}.
Formally, let $\mathcal{D}=\{(x_1,y_1), \dots, (x_n,y_n)\}$ be the training dataset. 
Then, the CAF objective is formulated as maximizing:
\begin{align}
\label{eqn:objective}
\sum_{i=1}^{n} \log \mathbb{P} \left[ F(x_i + \epsilon) = y_i \right] &= \sum_{i=1}^{n} \log \mathbb{E}  \left[ \mathbf{1} \left[ \arg\max F(x_i + \epsilon) = y_i \right] \right] \\ 
&= \sum_{i=1}^{n} \log \mathbb{E}  \left[ \mathbf{1} \left[ \arg\max f_{\theta_y}[f_{\theta_z}(x_i + \epsilon) + f_{\theta_a}(x_i + \epsilon)] = y_i \right] \right] \nonumber
\end{align}

Let $\{q_1,q_2,\dots,q_C\}$ be the output logits of $ f_{\theta_y}[f_{\theta_z}(x_i + \epsilon) + f_{\theta_a}(x_i + \epsilon)]$.
We use the softmax function to approximate the $\arg\max$ in Equation~\ref{eqn:objective} as a continuous and differentiable surrogate. 
Additionally, leveraging Jensen's inequality and the concavity of the logarithm function, we have the following lower bound:
\begin{equation*}
\sum_{i=1}^{n} \log \mathbb{E} \left[ \frac{e^{q_c}}{\sum_{c=1}^{C} e^{q_c}} \right] \geq  \sum_{i=1}^{n}  \mathbb{E} \left[ \log \frac{e^{q_c}}{\sum_{c=1}^{C} e^{q_c}} \right]
\end{equation*}
Therefore, by minimizing the cross-entropy loss, we improve the performance of the adaptive classifier $F(\cdot)$, maximizing the number of correctly classified noisy data samples in $\mathcal{D}$.
\subsection{Ensemble Adapters}
\begin{figure}
\centering
\includegraphics[width=0.95\columnwidth]{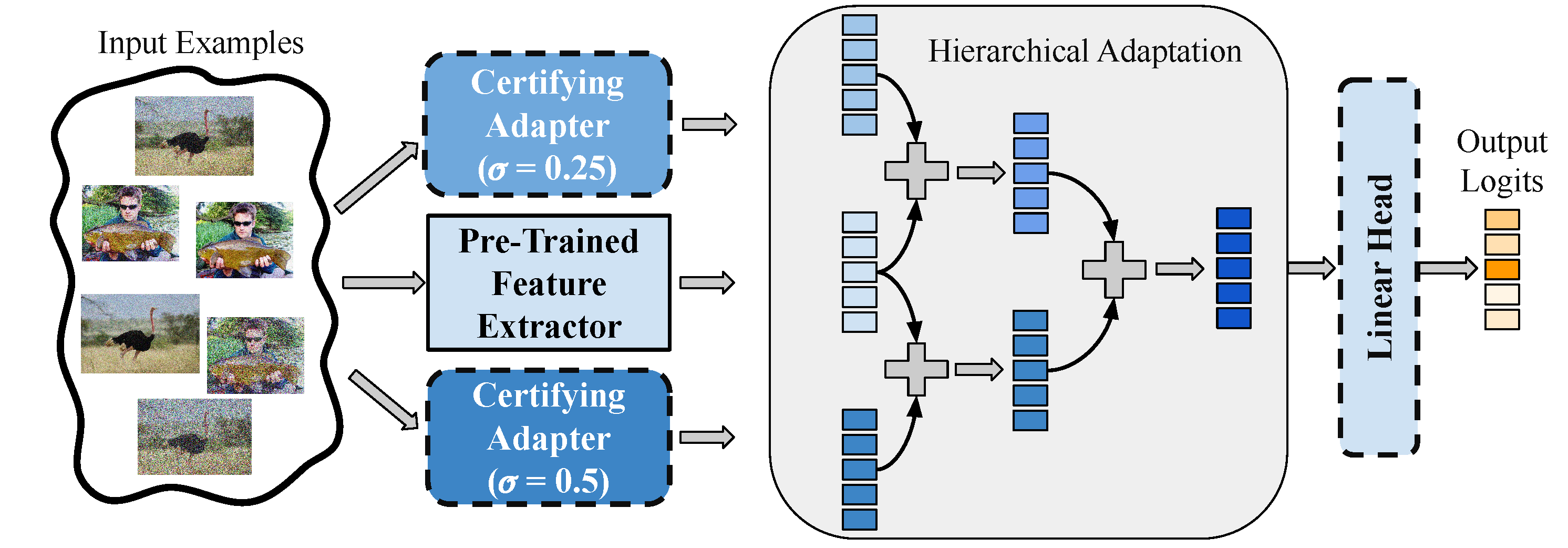}
\caption{\textbf{Ensemble of Certifying Adapters.}
A frozen pre-trained feature extractor is adapted to different noise scales (here, $0.25$ and $0.5$) with multiple certifying adapters through a hierarchical adaptation mechanism and a linear head.
Trainable model components are denoted with a dashed outline. 
}
\label{Fig:ensemble}
\end{figure}
Typically, achieving optimal certified robustness with methods such as RS~\cite{rs} and SmoothAdv~\cite{smoothadv} requires training several classifiers with the same architectures with different noise scales to match specific noise-augmentation distributions.
In contrast, CAF freezes the pre-trained feature extractor weights, but still requires the training of certifying adapters for each noise-augmentation distribution.
In the absence of prior knowledge concerning the scale of noise perturbations, we consider a variant of CAF where an ensemble of certifying adapters that are trained at different noise-augmentation distributions adapt a single feature extractor (Fig.~\ref{Fig:ensemble}).
A hierarchical adaptation scheme combines the latent representations of the certifying adapters and pre-trained feature extractors. 

Formally, let the set of trained certifying adapters be denoted $\{f^1_{\theta_a},\dots,f^V_{\theta_a}\}$; then, the forward process is formulated as:
\begin{equation*}
    f_{\theta_z}(x) + f_{\theta_a}(x) =
\sum_{v=1}^{V}(f_{\theta_z}(x) + f^i_{\theta_a}(x))
\end{equation*}
Since the following layer after the pre-trained feature extractor and certifying adapter is a linear layer, the above result can be further simplified as:
\begin{align*}
    f_{\theta_y} \left( \sum_{v=1}^{V}(f_{\theta_z}(x) + f^v_{\theta_a}(x)) \right) &= f_{\theta_y} \left( V\cdot f_{\theta_z}(x) + \sum_{v=1}^{V} f^v_{\theta_a}(x) \right) 
\end{align*}




\section{Empirical Results}
We evaluated CAF on two benchmark datasets (CIFAR-10 and ImageNet~\cite{5206848}) using both residual networks (ResNets)~\cite{He2015DeepRL} and ViTs~\cite{dosovitskiy2020vit}.
Following the same evaluation from RS~\cite{rs} and SmoothAdv~\cite{smoothadv}, we used Gaussian noise with $\ell_2$-norm perturbations and evaluated each method using certified accuracy, which is defined as the fraction of the test set images that are correctly classified by a smoothed classifier within an $\ell_2$ ball of radius $r$.
For the CIFAR-10 dataset, we evaluated robustness in $N = 100,000$ noise-augmented samples for each image in the test set. 
For the ImageNet dataset, we generated $N = 10,000$ noise-augmented samples from $500$ selected samples following RS~\cite{rs} to evaluate robustness. 
We primarily used the same noise augmentation methods for adapter training as described in RS~\cite{rs}, using three noise standard deviations, $\sigma \in \{0.25, 0.50, 1.00\}$. 
The batch size for the certifying adapters was set to $128$, except in the ensemble configuration where the batch size was set to $100$. 
For certifying adapters, we used a ResNet encoder (ResNet-18) for the pre-trained ResNet feature extractor and a low-rank adaptation ($Rank = 4$) approach~\cite{zhu2023melo} for the pre-trained ViT.
We used the ResNet-110 model provided by RS~\cite{rs} and SmoothAdv~\cite{smoothadv}, as well as the ViT (ViT-B/16 and ViT-L/16) pre-trained on ImageNet-1k. 
Both models were optimized using SGD with a learning rate of $0.01$.
We evaluated the ViT-B/16 and ViT-L/16 pre-trained on ImageNet-1k with the AdamW optimizer and a learning rate of $0.0001$~\cite{loshchilov2018decoupled}.
For competing methods, result tables report certified accuracies cited in their respective studies. 
All experiments were run on a server with an A100 GPU (40GB) and 8 CPU cores. 

\subsection{Classifier Adversarial Robustness}

\begin{figure}[h!]
    \centering
    \includegraphics[width=0.8\textwidth]{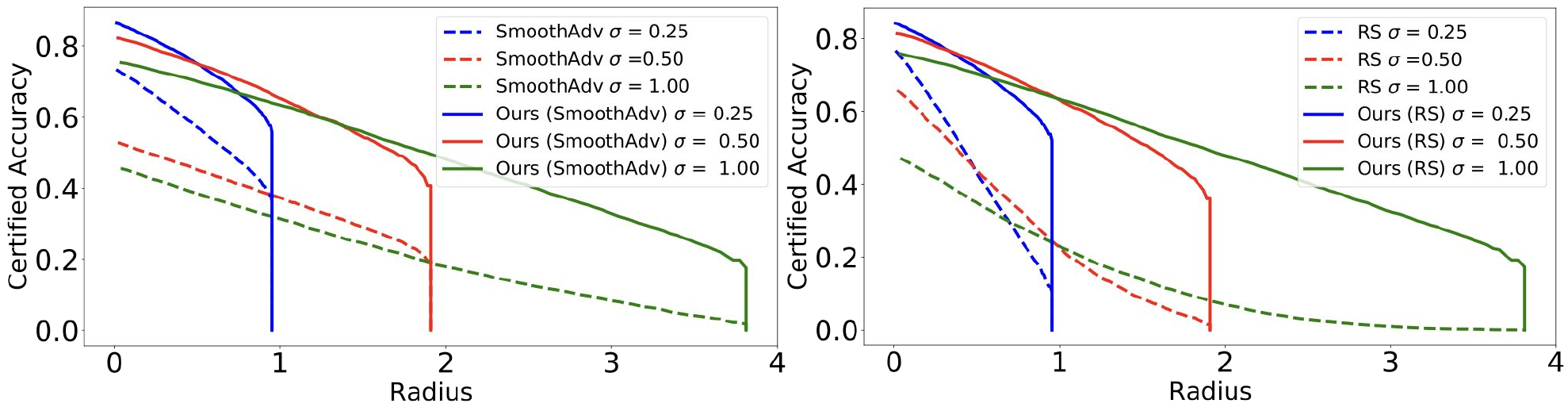} 
    \caption{\textbf{Certified accuracy across different radii.} We compared our method with RS~\cite{rs} and SmoothAdv~\cite{smoothadv}, using $3$ models trained at noise scales of 0.25, 0.50, and 1.00.}
    \label{fig:rs_smoothadv_comp}
\end{figure}

First, we compared CAF with randomized and denoised smoothing approaches in terms of certified accuracy across different radii in CIFAR-10 with ResNet feature extractors and certifying adapters (Table~\ref{tab:main_cifar10_resnet} and Fig.~\ref{fig:rs_smoothadv_comp}). 
We used the ResNet-110 model from RS~\cite{rs} and SmoothAdv~\cite{smoothadv} and incorporated a ResNet Encoder (the convolutional layers of a ResNet-18) as the certifying adapter for CAF.
Our method achieves state-of-the-art certified accuracy across selected radii in $[0,2.25]$ when compared to diffusion-based DDS~\cite{carlini2023certified} and DiffSmooth~\cite{diffsmooth}, optimization-based MACER~\cite{Zhai2020MACER}, SmoothMix~\cite{jeong2021smoothmix}, Consistency~\cite{10.5555/3495724.3496610}, and Boosting~\cite{2022boosting}, and random smoothing-based RS~\cite{rs} and SmoothAdv~\cite{smoothadv}.
The most notable improvements were observed at large radii, with CAF achieving a $2.31\times$ improvement over the second-best method (DiffSmooth$_{SmoothAdv}$) at a radius of $2.25$.

\begin{table}[h!]
\centering
\begin{tabular}{@{}ccccccccccc@{}}
\toprule
\multirow{2}{*}{Methods} & \multicolumn{10}{c}{Certified Accuracy (\%) under Radius $\textit{r}$}                                                                                                   \\ \cmidrule(l){2-11} 
                        & \textit{0.00} & \textit{0.25} & \textit{0.50} & \textit{0.75} & \textit{1.00} & \textit{1.25} & \textit{1.50} & \textit{1.75} & \textit{2.00} & \textit{2.25} \\ \midrule
RS~\cite{rs}                & 75.0          & 60.0          & 42.8          & 32.0          & 23.0          & 17.4          & 14.0          & 11.8          & 9.8           & 7.6           \\
SmoothAdv~\cite{smoothadv}               & 73.6          & 66.8          & 57.2          & 47.2          & 37.6          & 32.8          & 28.8          & 23.6          & 19.4          & 16.8          \\
MACER~\cite{Zhai2020MACER}                  & 81.0          & 71.0          & 59.0          & 47.0          & 38.8          & 33.0          & 29.0          & 23.0          & 19.0          & 17.0          \\
Consistency~\cite{10.5555/3495724.3496610}             & 77.8          & 68.8          & 58.1          & 48.5          & 37.8          & 33.9          & 29.9          & 25.2          & 19.5          & 17.3          \\
SmoothMix~\cite{jeong2021smoothmix}               & 77.1          & 67.9          & 57.9          & 47.7          & 37.2          & 31.7          & 25.7          & 20.2          & 17.2          & 14.7          \\
Boosting~\cite{2022boosting}                & 83.4          & 70.6          & 60.4          & 52.4          & 38.8          & 34.4          & 30.4          & 25.0          & 19.8          & 16.6          \\
DDS~\cite{carlini2023free}          & 79.0          & 62.0          & 45.8          & 32.6          & 25.0          & 17.6          & 11.0          & 6.20           & 4.20           & 2.20           \\
DiffSmooth$_{RS}$~\cite{diffsmooth}    & 78.2          & 67.2          & 59.2          & 47.0          & 37.4          & 31.0          & 25.0          & 19.0          & 16.4          & 14.2          \\
DiffSmooth$_{SmoothAdv}$~\cite{diffsmooth}   & 82.8          & 72.0          & 62.8          & 51.2          & 41.2          & 36.2          & 32.0          & 27.0          & 22.0          & 19.0          \\ \midrule
Ours$_{RS}$              & 84.2          & 78.6          & 73.2          & 68.1          & 62.0          & 59.6          & 55.1          & 51.6          & 47.6          & 43.8          \\
Ours$_{SmoothAdv}$        & \textbf{86.6} & \textbf{81.2} & \textbf{75.0} & \textbf{70.4} & \textbf{65.1} & \textbf{59.7} & \textbf{55.4} & \textbf{51.8} & \textbf{47.8} & \textbf{43.9} \\ \bottomrule
\end{tabular}
\caption{\textbf{Certified accuracy comparisons with other methods under radius $r$ with ResNet-110 on CIFAR-10 dataset.} DiffSmooth$_{RS}$ and DiffSmooth$_{SmoothAdv}$ indicate DiffSmooth method based on the certified classifier from RS~\cite{rs} and SmoothAdv~\cite{smoothadv}. Ours$_{RS}$ and Ours$_{SmoothAdv}$ indicate our proposed method with the certified classifier from RS~\cite{rs} and SmoothAdv~\cite{smoothadv}.}
\label{tab:main_cifar10_resnet}
\end{table}

Next, we evaluated CAF and randomized smoothing-based methods RS~\cite{rs} and SmoothAdv~\cite{smoothadv} at different noise scale and over multiple radii (Fig.~\ref{fig:rs_smoothadv_comp}).
We used three ResNet-110 models, each trained at noise scales of $0.25, 0.50$, and $1.00$ to compute certified accuracies. 
Across all noise scales, CAF outperformed RS and SmoothAdv, with the improvement becoming more pronounced at larger radius (consistent with the general trend of Table~\ref{tab:main_cifar10_resnet}). 

To demonstrate that CAF can certify robustness for models pre-trained on benchmark (clean) data, we considered a ViT-B/16 feature extractor pre-trained on CIFAR-10 and ViT-L and ViT-B feature extractors pre-trained on ImageNet-1K.
We evaluated CAF, DMAE~\cite{wu2023dmae}, and DDS~\cite{carlini2023certified} on CIFAR-10 (Table~\ref{tab:main_cifar10_vit}).
In addition to these methods, we also evaluated RS~\cite{rs} and DiffSmooth~\cite{diffsmooth} on ImageNet (Table~\ref{tab:main_imgnet_vit}).
Our method achieved the leading certified accuracy across all radii in CIFAR-10, but the results were less consistent in ImageNet.
Diffusion-based models out-performed CAF at radii of $0.5$, $2.0$, and $3.0$. though CAF achieved a higher certified accuracy compared to non-diffusion-based models.
Given the performance of both diffusion and adapter-based approaches, these results suggest that a combined diffusion-adapter based robustness certification method may both perform well empirically and effectively scale to ImageNet data. 


\begin{table}[h!]
\centering
\begin{tabular}{@{}llllll@{}}
\toprule
\multirow{2}{*}{Architecture} & \multirow{2}{*}{Methods}      & \multicolumn{4}{l}{Certified Accuracy (\%) at $\textit{r}$} \\ \cmidrule(l){3-6} 
                              &                               & \textit{0.25}   & \textit{0.50}   & \textit{0.75}   & \textit{1.00}   \\ \midrule
\multirow{7}{*}{ViT-B/16}     & DMAE~\cite{wu2023dmae}                          & 79.2            & 64.6            & 47.3            & 36.1            \\
                              & DMAE + continued pre-training~\cite{wu2023dmae} & 80.8            & 67.1            & 49.7            & 37.7            \\
                              & DDS~\cite{carlini2023certified}                           & 76.7            & 63.0            & 45.3            & 32.1            \\
                              & DDS + fine-tuning~\cite{carlini2023certified}                & 79.3            & 65.5            & 48.7            & 35.5            \\
                              & Ours                          & \textbf{97.1}   & \textbf{96.0}   & \textbf{94.6}   & \textbf{92.8}   \\ \bottomrule
\end{tabular}
\caption{\textbf{Certified accuracy comparisons with DDS~\cite{carlini2023certified} and DMAE~\cite{wu2023dmae} under radius $r$ using ViT-B/16 on CIFAR-10.} We also evaluated with standard DMAE + continued pre-training and DDS + fine-tuning.}
\label{tab:main_cifar10_vit}
\end{table}

\begin{table}[h!]
\centering
\begin{tabular}{@{}cccccccc@{}}
\toprule
\multirow{2}{*}{Methods}                               & \multirow{2}{*}{\begin{tabular}[c]{@{}c@{}}Classifier\\ Architecture\end{tabular}} & \multirow{2}{*}{\#Params} & \multicolumn{5}{c}{Certified Accuracy (\%) at $\textit{r}$}                            \\ \cmidrule(l){4-8} 
                                                      &                                                                                    &                           & \textit{0.5}  & \textit{1.0}  & \textit{1.5}  & \textit{2.0}  & \textit{3.0}  \\ \midrule
RS~\cite{rs}                    & ViT-L                                                                              & 304.4M                    & 70.2          & 51.8          & 38.4          & 32.0          & 17.0          \\
DDS~\cite{carlini2023certified} & ViT-L                                                                              & 857M                      & 71.1          & 54.3          & 38.1          & 29.5          & 13.1          \\
DiffSmooth~\cite{diffsmooth}    & ViT-L                                                                              & 857M                      & \textbf{77.2} & 63.2          & 53.0          & 37.6          & 24.8          \\
DMAE~\cite{wu2023dmae}         & ViT-B                                                                              & 86.5M                     & 69.6          & 57.9          & 47.8          & 35.4          & 22.5          \\
DMAE~\cite{wu2023dmae}          & ViT-L                                                                              & 304.3M                    & 73.6          & 64.6          & 53.7          & \textbf{41.5} & \textbf{27.5} \\ \midrule
Ours                                                  & ViT-B                                                                              & 86.7M                     & 71.8          & 53.6          & 45.8          & 34.2          & 21.2          \\
Ours                                                  & ViT-L                                                                              & 304.7M                    & 75.4          & \textbf{65.1} & \textbf{54.3} & 36.7          & 23.9          \\ \bottomrule
\end{tabular}
\caption{\textbf{Certified accuracy comparisons with RS~\cite{rs}, DDS~\cite{carlini2023free}, DiffSmooth~\cite{diffsmooth} and DMAE~\cite{wu2023dmae} under radius $r$ with ViTs on ImageNet.} 
The ViT-L used in RS~\cite{rs}, DDS~\cite{carlini2023free}, and DiffSmooth~\cite{diffsmooth} is pre-trained in a self-supervised manner.  }
\label{tab:main_imgnet_vit}
\end{table}

\subsection{Adaption with Ensemble Adapters}

We next evaluated the certified accuracy of the proposed ensemble CAF using ViT-B/16 on the CIFAR-10 and ImageNet datasets. 
Three certifying adapters were trained in single CAF models using a noise scale  $\sigma \in \{0.25, 0.50, 1.00\}$.
These adapters were then combined with a pre-trained feature extractor into the ensemble CAF using our hierarchical adaptation scheme.
We fine-tuned the certifying adapters and linear head using training samples that were randomly perturbed using noise scales $\sigma \in \{0.25, 0.50, 1.00\}$.  
To assess performance, we uniformly sampled 500 images from the ImageNet test set and 1,000 images from the CIFAR-10 test set. 
Finally, we generated 10,000 noised examples for each selected image to evaluate certified accuracy. 

\begin{figure}[h!]
    \centering
    \includegraphics[width=0.7\textwidth]{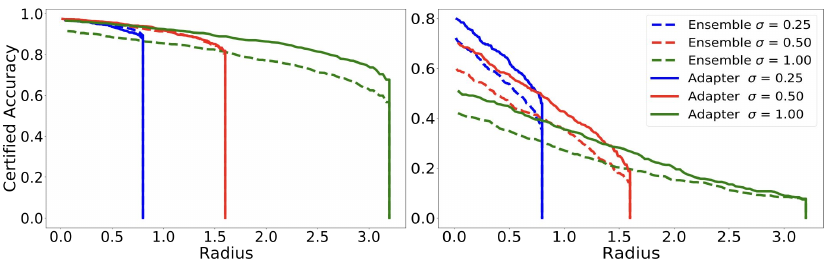} 
    \caption{\textbf{Certified accuracy across different radii for the single adapter CAF and ensemble CAF.} 
    Each CAF configuration was trained on CIFAR-10 (left) and ImageNet (right) using a ViT-B/16. 
    Singular adapters are trained and evaluated at the same noise scale. 
    The ensemble adapters are trained using mixed noise scales and assessed for certified accuracy using   noise scales: 0.25, 0.50, and 1.00. }
    \label{fig:ensemble_vs_single}
\end{figure}

First, we compared the certified accuracies of the ensemble CAF with the single-adapter CAF using the same noise scale for evaluation and training (Fig.~\ref{fig:ensemble_vs_single}).
For the CIFAR-10 dataset, the performance of the ensemble adapters closely matches that of the individual adapter at noise scales of $\sigma = 0.25$ and $\sigma = 0.50$, and is marginally better at $\sigma = 0.25$ and $r \in (0.50,0.75)$. 
However, the individual adapter outperforms the ensemble at the larger noise scale of 1.00. 
On the ImageNet dataset, the individual adapter consistently outperforms the ensemble adapters across all three noise scales.
This behavior is not too surprising given that the single certifying adapter configuration assumed the same noise scale in both the training and evaluation.

We further considered the scenario where the single adapter may not assume the same noise scale in training and evaluation in both CIFAR-10 and ImageNet datasets (Table~\ref{tab:ensemble_vs_single}).
Since the single adapter CAF configurations were trained with specific noise scales, they outperform other single adapters and the ensemble when evaluating at the same noise scale. 
However, the ensemble adapters showed balanced performance across various noise scales, despite exhibiting slightly lower classification accuracy compared to the individual adapters for their corresponding training noise scales. 
Training a single adapter with data sampled at different noise scales performed considerably worse than the ensemble CAF (Table~\ref{tab:ensemble_vs_single}, Adapter (Mix)). 
Considering average classification accuracy, the ensemble adapters achieved the best average classification accuracy across all settings.

Our proposed ensemble adapters exhibited robustness across multiple noise scale perturbations and achieved certified accuracy under various noise conditions compared to individual adapter structures. 
This suggests that ensemble adapters might offer a better solution to practical challenges in real-world scenarios, specifically in achieving \textit{optimal performance when the noise scale is unknown.}
\begin{table}[h!]
\centering
\resizebox{0.9\columnwidth}{!}{%
\begin{tabular}{@{}c|cccc|cccc@{}}
\toprule
Datasets                & \multicolumn{4}{c|}{CIFAR-10}                                     & \multicolumn{4}{c}{ImageNet}                                      \\ \midrule
Testing $\sigma$             & 0.25           & 0.50           & 1.00           & Avg.           & 0.25           & 0.50           & 1.00           & Avg.           \\ \midrule
Adapter (Training $\sigma$=0.25) & \textbf{98.20} & 97.05          & 81.75          & 92.33          & \textbf{75.81} & 37.46          & 0.23           & 37.83          \\
Adapter (Training $\sigma$=0.50) & 97.10          & \textbf{97.88} & 91.20           & 95.39          & 55.65          & \textbf{65.45} & 0.21           & 40.44          \\
Adapter (Training $\sigma$=1.00) & 96.32          & 96.27          & \textbf{96.83} & 96.47          & 29.10          & 28.81          & \textbf{45.98} & 34.63          \\
Adapter (Mix)           & 94.94          & 94.05          & 93.98          & 94.32          & 55.31          & 55.14          & 16.29          & 42.25          \\
Ensemble Adapters       & 97.60          & 97.44          & 96.42          & \textbf{97.15} & 73.32          & 62.82          & 41.67          & \textbf{59.27} \\ \bottomrule
\end{tabular}
}
\caption{\textbf{Ensemble certified accuracy.}
The four single certifying adapter configurations (Adapter) and ensemble CAF configuration (Ensemble Adapters) were evaluated under different noise scale perturbations on the full CIFAR-10 and ImageNet test dataset using ViT-B/16.}
\label{tab:ensemble_vs_single}
\end{table}
\subsection{CAF Sensitivity of Certifying Adapter Parameters}

\begin{figure}[h!]
    \centering
    \includegraphics[width=1\textwidth]{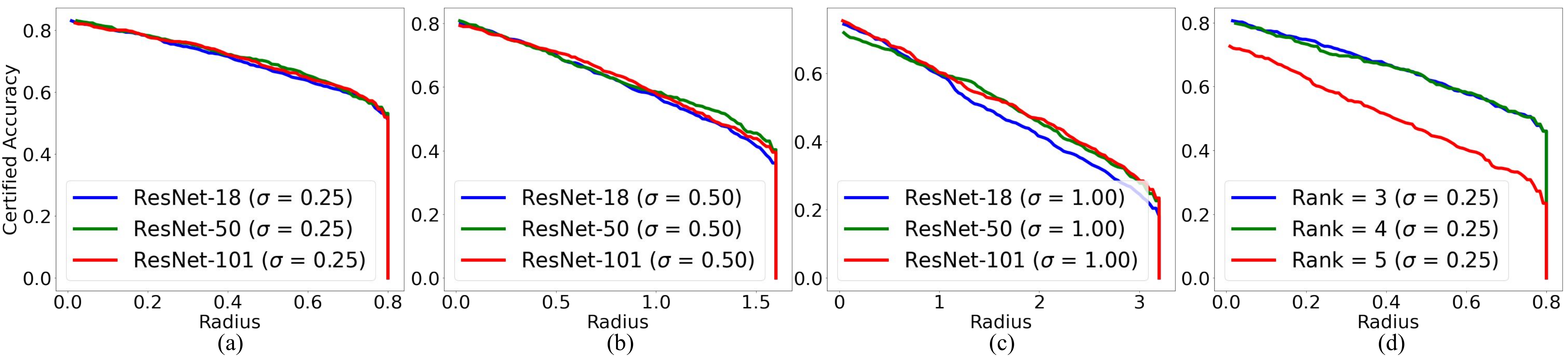} 
    \caption{\textbf{Certified accuracy with different size adapters.} 
    The certified accuracies over radii in $[0.0,0.8]$ are shown for noise perturbations with scales (a) $\sigma = 0.25$, (b) $\sigma = 0.50$, and (c) $\sigma = 1.00$ for different sized ResNet encoders as adapters on CIFAR-10. 
    (d) Certified accuracy with a varied certifying adapter rank for ViT-B/16 on ImageNet ($\sigma = 0.25$).}
    \label{fig:ablation_study}
\end{figure}

Finally, we investigated how changing the configurations of the adapters would affect the certified accuracy of CAF.
We used three ResNet encoders (ResNet-18, ResNet-50, and ResNet-101) and three noise scales (${0.25, 0.50, 1.00}$) to evaluate CAF on CIFAR-10. 
We observed that adding more convolutional layers does not consistently or noticeably impact the performance of certified accuracy (Fig.~\ref{fig:ablation_study}). 
The ResNet-50 and ResNet-101 certifying adapters contributed to a marginal improvement in the certified accuracy of CAF for radii greater than 1 (Fig.~\ref{fig:ablation_study} (c)), but minimal improvements at a noise scale of $0.25$ or $0.50$ (Fig.~\ref{fig:ablation_study} (a,b)).
We further evaluated the impact of varying the adapter rank for ViT-B/16 (Fig.~\ref{fig:ablation_study} (d)).
Interestingly, rank $3$ and $4$ adapters showed substantial improvement over the larger rank $5$ adapter, suggesting that a higher rank does not guarantee coverage of more meaningful spaces, echoing the experimental observations found in LoRA~\cite{hu2022lora}. 
This implies that a low-rank adaptation matrix is sufficient for achieving competitive certified accuracy.
These sensitivity results suggest that CAF is relatively insensitive to the certifying adapter parameters. 

\section{Discussion}

One particular benefit of CAF is its ability to adapt to different pre-trained models, without having to perform the resource-intensive task of training from scratch. 
Training large models like ViTs or ResNets on the ImageNet dataset typically takes several days, even when using a 40GB A100 GPU. 
Likewise, it is resource intensive to achieve certified robustness using randomized smoothing as it requires training multiple models each with a different noise scale~\cite{rs,smoothadv}. 
When starting with a pre-trained classifier, full model retraining is generally preferred to fine-tuning only the head to ensure the model is sufficiently adapted to new tasks or data distributions; but, computing full model updates can be computationally prohibitive. 
The CAF framework overcomes the aforementioned issues by leveraging the feature extraction capabilities of either certified or uncertified pre-trained models, avoiding expensive training regimes.



In terms of applicability and limitations, CAF does require white-box access to the pre-trained feature extractor, whereas randomized smoothing only requires class predictions~\cite{rs}. 
Additionally, the single certifying adapter CAF configurations assume the noise scale at testing is the same as the noise scale used for training (similar to randomized smoothing). 
However, we showed that this issue can be mitigated in CAF by using an ensemble of certifying adapters.


\section{Conclusions}

In this work, we introduced CAF, a certifying adapters framework which enables and enhances the certification of adversarial robustness for classifiers.
We showed that certifying adapters can certify uncertified models (e.g., ViT-B/16 and ViT-L/16) pre-trained on clean datasets and substantially improve the performance of state-of-the-art certified models (ResNet-110) via randomized smoothing~\cite{rs} and SmoothAdv~\cite{smoothadv} at multiple radii in CIFAR-10 and ImageNet. 
The most notable improvements in certified accuracy were observed for models trained with RS and SmoothAdv; e.g., 5.76$\times$ and 2.61$\times$ improvements for $r=2.25$ in CIFAR-10 when compared to RS and SmoothAdv, respectively. 
We demonstrated that an ensemble of adapters enables a single pre-trained feature extractor to defend against a range of noise perturbation scales.
Finally, we showed that CAF is insensitive to adapters of different sizes, indicating that the effectiveness of certifying adapters may arise from the framework itself, rather than from adding additional model complexity.

\clearpage
\bibliographystyle{IEEEtranS} 
\bibliography{ref}

\end{document}